\documentclass[sigconf]{acmart}
\AtBeginDocument{%
  \providecommand\BibTeX{{%
    \normalfont B\kern-0.5em{\scshape i\kern-0.25em b}\kern-0.8em\TeX}}}

\usepackage{multirow}
\usepackage{balance}

\copyrightyear{2021}
\acmYear{2021}
\setcopyright{acmcopyright}
\acmConference[MM '21] {Proceedings of the 29th ACM Int'l Conference on Multimedia}{October 20--24, 2021}{Virtual Event, China.}
\acmBooktitle{Proceedings of the 29th ACM Int'l Conference on Multimedia (MM '21), Oct. 20--24, 2021, Virtual Event, China}
\acmPrice{15.00}
\acmISBN{978-1-4503-8651-7/21/10}
\acmDOI{10.1145/3474085.3475318}

\begin{document}
\fancyhead{}

\title{Towards Realistic Visual Dubbing with Heterogeneous Sources}


\author{Tianyi Xie}
\affiliation{
    \institution{Shanghai Jiao Tong University}
    \country{}
}
\email{tianyixie77@gmail.com}

\author{Liucheng Liao}
\affiliation{
    \institution{University of Electronic Science and Technology of China}
    \country{}
}
\email{liuchengliao@gmail.com}

\author{Cheng Bi}

\author{Benlai Tang}
\authornote{Corresponding author.}

\author{Xiang Yin}
\affiliation{
    \institution{ByteDance AI Lab}
    \country{}
}
\email{tangbenlai@bytedance.com}

\author{Jianfei Yang}
\affiliation{
    \institution{Nanyang Technological University}
    \country{}
}
\email{yang0478@e.ntu.edu.sg}

\author{Mingjie Wang}
\affiliation{
    \institution{University of Guelph}
    \country{}
}
\affiliation{
    \institution{Memorial University of Newfoundland}
    \country{}
}

\author{Jiali Yao}
\author{Yang Zhang}
\author{Zejun Ma}
\affiliation{
    \institution{ByteDance AI Lab}
    \country{}
}




\begin{abstract}
The task of few-shot visual dubbing focuses on synchronizing the lip movements with arbitrary speech input for any talking head video. Albeit moderate improvements in current approaches, they commonly require high-quality homologous data sources of videos and audios, thus causing the failure to leverage heterogeneous data sufficiently. In practice, it may be intractable to collect the perfect homologous data in some cases, for example, audio-corrupted or picture-blurry videos. To explore this kind of data and support high-fidelity few-shot visual dubbing, in this paper, we novelly propose a simple yet efficient two-stage framework with a higher flexibility of mining heterogeneous data. Specifically, our two-stage paradigm employs facial landmarks as intermediate prior of latent representations and disentangles the lip movements prediction from the core task of realistic talking head generation. By this means, our method makes it possible to independently utilize the training corpus for two-stage sub-networks using more available heterogeneous data easily acquired. Besides, thanks to the disentanglement, our framework allows a further fine-tuning for a given talking head, thereby leading to better speaker-identity preserving in the final synthesized results. Moreover, the proposed method can also transfer appearance features from others to the target speaker. Extensive experimental results demonstrate the superiority of our proposed method in generating highly realistic videos synchronized with the speech over the state-of-the-art. 
\end{abstract}

\ccsdesc[500]{Computing methodologies~Computer vision}
\ccsdesc[500]{Computing methodologies~Phonology / morphology}

\keywords{few-shot, visual dubbing, heterogeneous sources, two stages}

\begin{teaserfigure}
  \includegraphics[width=\textwidth]{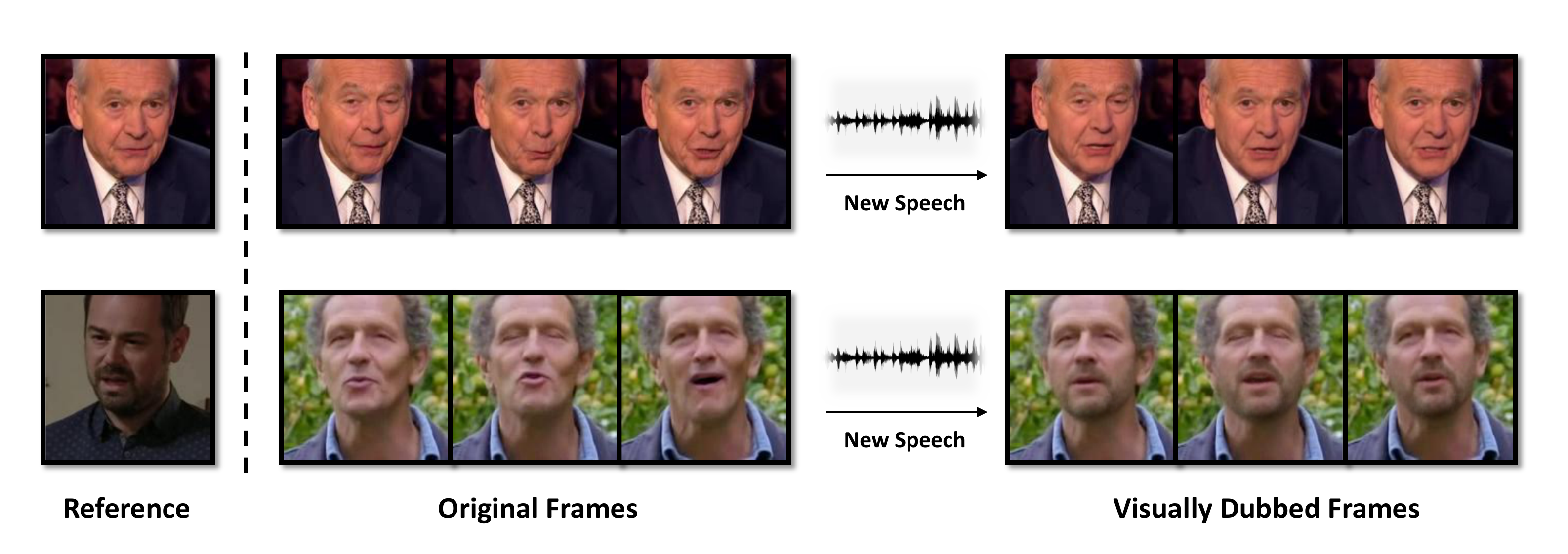}
  \caption{The examples provided by our framework, which supports visual dubbing for any talking head videos with arbitrary speech input. With a reference talking head fed into, our method is also capable of transferring its lower-half face representations to the target speaker.}
  \label{fig:head_image}
\end{teaserfigure}

\maketitle

\section{Introduction}
Due to the presence of language gaps, people always have trouble in understanding when watching foreign movies or news without localized subtitles. To make media content cater for local viewers, video producers have made a lot attempts to replace the original foreign language with the native language, namely, dubbing. However, such audio-only dubbing inevitably gives rise to mismatch between visual signal and audio content. To be specific, the speaker's lip movements do not synchronize with his/her corresponding voice, which leads to daunting experience in understanding the speech for viewers, especially for the hearing-impaired ones heavily relying on lip-reading. Therefore, it is highly expected to develop visual dubbing methods, which could synchronize the visual signal with the replaced new audio channel.

Based on the difference of generalization, previous related works could be roughly classified into two categories. The first, namely,  personalized facial reenactment task \cite{suwajanakorn2017synthesizing, fried2019text, afouras2018deep}, tries to synthesize realistic talking head videos given a driving source (e.g. a piece of audio input or a sequence of facial landmarks). To generate realistic talking heads, \cite{kim2018deep} considers the facial landmarks of the source actor as input, whereas \cite{fried2019text} uses text as the driving source. Although some promising and vivid results have been achieved by these works, they usually require hours of video footage for a single speaker, which is tedious and difficult to access and collect in practical applications. To alleviate this issue, the other kind of task tries to synthesize realistic talking heads by using sets of facial images of a certain person even if this person is unseen in the training set. In spite of robustness produced by existing approaches \cite{zakharov2019few, ha2019marionette, chen2020talking}, severe visual artifacts emerge in the generated results (especially in the background regions) because of data deficiency of the target speaker in the few-shot task.

To further tackle these issues, a series of approaches~\cite{kr2019towards, jamaludin2019you, prajwal2020lip} are present to morph the lower face area according to the audio input rather than synthesizing the whole image in the task of talking head generation, which is named as few-shot lip-synchronization or few-shot visual dubbing. Specifically, they successfully learn a non-linear projection mapping from speech representation to the lip movements and allow visual dubbing for unconstrained videos with arbitrary speech. However, the following challenges still remain in current few-shot visual dubbing problems:\emph{(i)} Almost all of existing methods are trained in an end-to-end manner and thus put a burden on the requirement of high-quality pairwise data of audio and video, which is difficult and costly to collect. Consequently, heterogeneous sources, such as videos with background music and low-resolution videos, cannot be fully made use of for their training to advance the generation quality, thus leading to the waste of data; \emph{(ii)} The visual quality of generated images is still far from realistic as expected. In particular, awful blurry mouth and teeth usually occur in the synthesized results; \emph{(iii)} Identity gaps between the generated and input videos can not be further eliminated, which is undesirable in the real-world applications of visual dubbing.

It is widely known that deep neural networks are basically data-driven and big training sets always contribute to the preeminent performance and convergence of the networks. Hence, it is intuitive and promising to design a method that is able to fully make use of heterogeneous data, with the purpose of enlarging the training set. For example, GIF files without audio is potential for guiding realistic image generation while blurry black-and-white videos can provide additional clues for the mapping between speech signals and lip movements. However, the existing end-to-end frameworks are incapable of utilizing these data due to the low quality or even lack of either audios or pictures. Moreover, the issue of identity gap is also intractable for previous systems \cite{kr2019towards,prajwal2020lip} in the few-shot setting. A further fine-tuning on a specific speaker may help narrow this gap, but it is difficult to be incorporated into previous frameworks caused by the nature of their architectures. In this case, the models may be prone to overfit to the particular speaker's speech and cannot well generalize to others' anymore.

To overcome above-mentioned challenging issues, we propose a novel framework to generate vivid visually-dubbed videos in a two-stage manner that ensures a more flexibility of using data. Specifically, the first stage is audio-related and is designed to generate lip-synced landmarks on top of pose prior and audio input, whereas the second, image-related stage, is dedicated to translate landmarks to realistic faces. It is worth noting that both of these two stages are trained in a few-shot fashion. Benefiting from the two-stage paradigm, it is feasible to disentangle the lip movements prediction from the phase of realistic image generation instead of direct audio-visual mapping. In this way, training corpora for each stage can be separable and independent. One potential benefit is that we can use talking head videos of low-resolution to train the sub-network of audio-related landmark generation while movie clips accompanied by background music for the training of landmark translation stage. Additionally, for any given specific speakers, we can only fine-tune the landmark translation stage to better capture personal appearance while the audio-related sub-network still allows the inputs of arbitrary voice. Our core contributions are summarized as follows:
\begin{itemize}
    \item A novel few-shot neural visual dubbing framework is proposed, which consists of two stages and has the capacity of editing video with arbitrary speech input. The entire canonical task is subtly divided into two separate few-shot sub-tasks, which not only achieves the flexibility of using heterogeneous data, but also enhances the model to generate vivid visually dubbed videos with well identity preserving. 
    \item A few-shot landmark generator is present to predict lip-synced landmarks on the basis of acoustic features and acoustic-unrelated landmarks, which attenuates the challenging issue of generating lower-half faces directly from acoustic features in the image generator. 
    \item We design a new generative model that can synthesize highly realistic lower-half faces under the few-shot settings, which can be put into original video footage seamlessly. The incorporation of Adaptive Instance Normalization (AdaIN) \cite{karras2019style} enables the model to transfer anyone's lower-half facial representations to the synthesized results.
    \item Extensive quantitative experiments and subjective evaluations are conducted to compare our framework with several state-of-the-art approaches, which shows that our proposed method achieves significant improvement in visual quality, identity similarity with competitive audio-lip consistency.
\end{itemize}

\section{Related Work}
\subsection{Personalized facial reenactment}
Facial reenactment changes target poses and expressions based on the driving source. Generally, the source can be categorized as performance-driven, speech-driven, and text-driven. For performance-driven facial reenactment, \cite{kim2018deep} allows a source actor to fully control the motion of the target actor, like a marionette. Moreover, \cite{kim2019neural} preserves the speaking style of the target actor when performing control through training a cycle GAN model \cite{zhu2017unpaired} in the expression subspace, which can translate expression style from the source actor to the target. \cite{yang2020large} also describes a system for large-scale audiovisual translation and dubbing. Note that the landmark or blend shape is usually employed as the intermediate product in the facial reenactment task. As to speech-driven or text-driven tasks, recent works \cite{yu2019mining, fried2019text, suwajanakorn2017synthesizing, thies2019neural} learn a mapping from driving source to the intermediate result and then performs a conditional image-to-image translation task. Thanks to advances in the image translation field \cite{isola2017image, wang2018high}, realistic generation can be achieved. However, this line of works usually requires a large amount of video footage of a particular speaker, typically a few hours. In contrast, our work generalizes to any identity and works for arbitrary speech input.

\begin{figure*}[t]
\begin{center}
\includegraphics[width=1.0\linewidth]{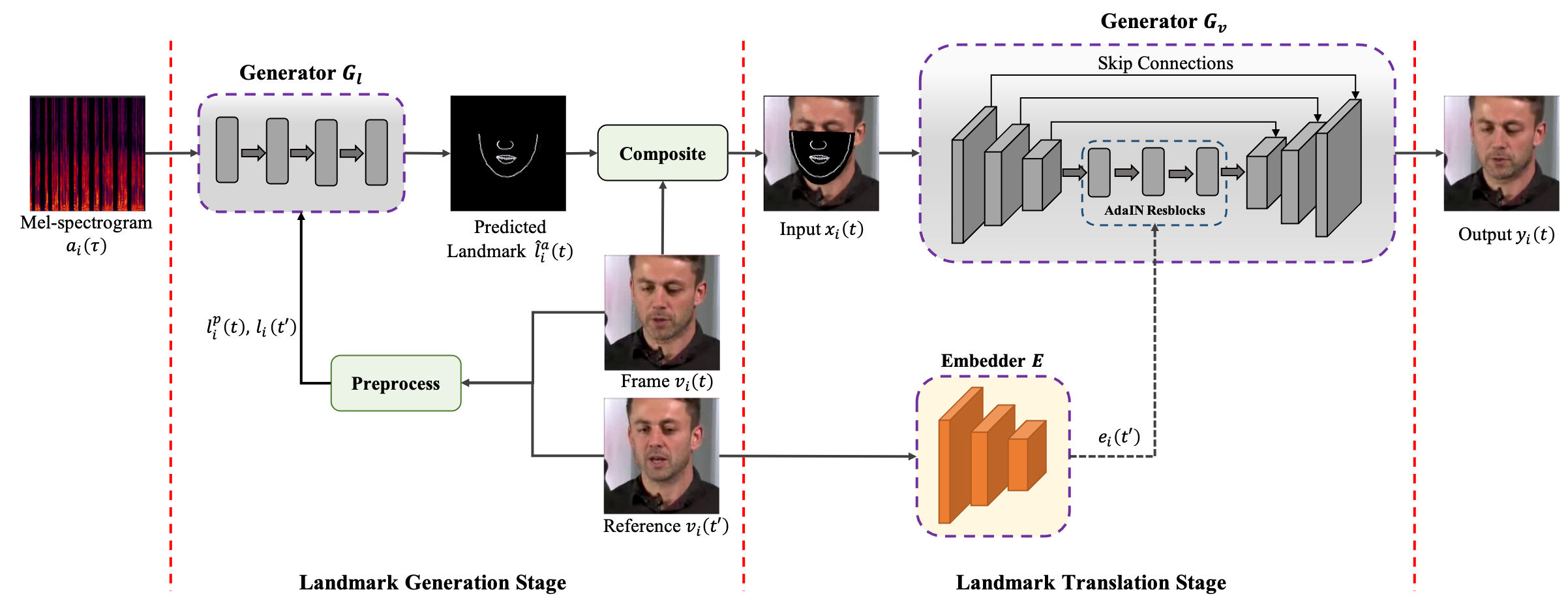}
\end{center}
   \caption{For more flexible use of data, we introduce a two-stage framework that consists of the landmark generation stage and the landmark translation stage. With this design, we decouple the original visual dubbing task into lip movements prediction and realistic image generation, which enables us to make use of unpaired data and generate realistic visual dubbing results.}
\label{fig:model overview}
\end{figure*}

\subsection{Few-shot talking head generation}
\cite{chen2020comprises} formulates the problem of few-shot talking head generation as follows: \textit{given one (or a few) image(s) and a driving source(e.g. a piece of audio speech or a sequence of facial landmarks), the task of talking-head video generation is to synthesize realistic-looking, animated talking-head video that corresponds to the driving source.} \cite{zakharov2019few} proposed a meta-learning framework with AdaIN \cite{huang2017arbitrary} technique to solve the few-shot image translation issue. \cite{ha2019marionette} utilizes an image attention block to better preserve target identity and also introduces a landmark transformer used to adapt source landmarks to the target one. \cite{chen2019hierarchical, zhou2020makeittalk} successfully animate a single facial image based on audio input while \cite{chen2020talking} further delves into the relationship between the speech and head motion. Besides, \cite{zakharov2020fast} makes real-time talking head generation possible in the few-shot setting. However, despite their breakthrough, the unnaturalness in the generated results is still visible, given the acuity of human perceptions. Hence, at present, full talking head generation of any identity can not be put into practical use yet.

\subsection{Few-shot lip synchronization}
Different from talking head generation, lip synchronization, essentially the same task as visual dubbing, is more like a video editing task, which aims to morph the lip movements instead of generating the whole image. Though several works \cite{thies2019neural, suwajanakorn2017synthesizing} show verisimilar lip-synced results, they are still limited to the particular speaker. As far as we know, very few works have been designed to solve lip synchronization for any identity and voice. Initial works \cite{jamaludin2019you, kr2019towards} achieve lip-sync results when animating a static image but can not work for a dynamic sequence, such as a video. Very recently, utilizing a strong lip-sync discriminator, \cite{prajwal2020lip} first proposed a speaker-independent model to synchronize the lip movements of unconstrained videos with arbitrary speech input. Similar to \cite{kr2019towards}, \cite{prajwal2020lip} also takes target face with bottom-half masked as input. Still, they are unable to utilize unpaired data, which may be unsatisfactory in audio or picture, and there is notable distinction between their synthesized results and real videos. The blur of mouth and teeth remains a common problem in their results. To address these issues, we build a two-stage framework to decouple the overall task and each stage performs an independent few-shot task. Such a two-stage design makes it possible to leverage unpaired data and enables us to generate more realistic lip-sync videos.

\section{Methods}

\subsection{Overview}
We introduce a two-stage approach for few-shot visual dubbing, which takes video frames $V^T_i=\{v_i(1), v_i(2), ..., v_i(T)\}$, and a sequence of acoustic features $A^T_i=\{a_i(1), a_i(2), ..., a_i(cT)\}(c>1)$, as input and synthesized audio-visual synced video frames, $Y^T_i=\{y_i(1), y_i(2), ..., y_i(T)\}$. Here, $v_i$ denotes the $i$-th video and $v_i(t)$ represents the $t$-th frame of this video. The same is with acoustic features but its temporal resolution is higher than visual features. To decouple lip-sync and realistic generation, we divide the whole process into two stages (Figure \ref{fig:model overview}) and use a landmark with 216 2D facial keypoints as the intermediate representation. Training of each stage is separate and independent and the adopted landmark detection method is basically borrowed from \cite{wu2018look}. The first stage, few-shot landmark generation, mainly works on landmarks space instead of complex pixel space and adjust lower-half landmark $l^a_i(t)$ to be synchronized with speech. Since we only need to modify the lower-half landmark, the upper part $l^p_i(t)$, including the nose, eyes, and eyebrows, can be used as the pose prior. In addition to speech and upper-part landmarks, we also input a reference landmark $l_i(t')$ to help the model capture person-specific features of lower-half landmarks. Note that we set $t \neq t'$ in training while usually $t = t'$ in inference. Then, the second stage, few-shot landmark translation, performs an image-to-image translation task to paint the modified landmark area conditioned on the processed composite input $x_i(t)$. Similarly, a reference frame $v_i(t')$ is also used for providing personalized facial features. Inspired by \cite{zakharov2019few}, our few-shot landmark translation adopts a meta-learning approach and utilizes AdaIN \cite{karras2019style} method to incorporate the reference appearance feature. After the meta-training phase converges, fine-tuning with the given speaker enables the model to better capture his/her appearance details and bridge the identity gap between synthesized and real talking heads. The detailed implementation can be found in the supplementary material.

\subsection{Few-shot Landmark Generation}
The landmark generator $G_l$ aims to synthesize lip-synced landmarks conditioned on acoustic features $a_i(\tau)$ and pose landmarks $l^p_i(t)$. However, utilizing only $a_i(\tau)$ and $l^p_i(t)$ are insufficient for landmarks prediction since they lack appearance information because the model tends to predict average landmark for any people unseen. Therefore, we add reference landmark $l_i(t')$ as a supplementary condition. The function of landmark generator $G_l$ can be formulated as follows:
\begin{equation}
    l^a_i(t) = G_l(a_i(\tau), l^p_i(t), l_i(t'))
\end{equation}

\subsubsection{Network Architecture.} 
\begin{figure}[t]
\begin{center}
   \includegraphics[width=1.0\linewidth]{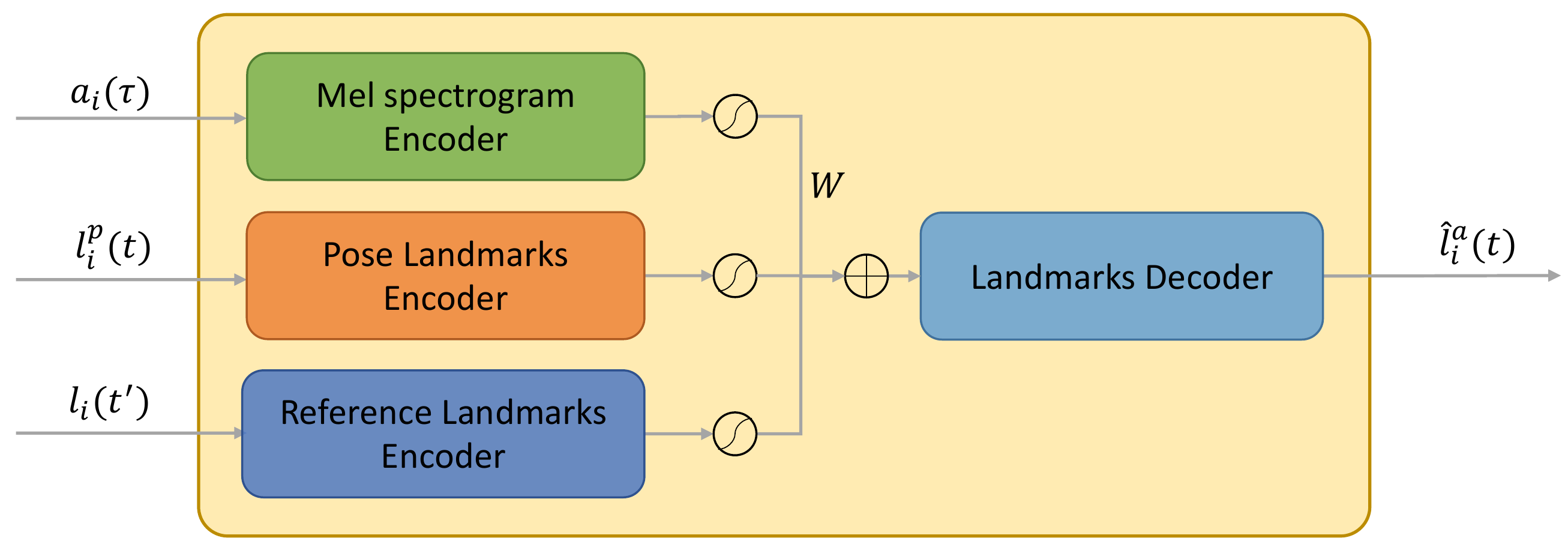}
\end{center}
   \caption{The landmark generator $G_l$ consists of three encoders and one decoder.}
\label{fig:landmark generator}
\end{figure}

The generator $G_l$ consists of three encoders and one decoder (see Figure \ref{fig:landmark generator}). Similar to \cite{prajwal2020lip}, our mel-spectrogram encoder consists of fully convolution residual blocks. We extract 80-band mel-spectrogram as $a_i(\tau)$. The frame window size and frame shift are set to 40ms and 10ms respectively. We set $\tau$ to $5t$ so that each temporal mel-spectragrams represent 200ms. Besides, we add the reference landmark as a global embedding instead of a frame-level local condition. More precisely, we sample all frames landmarks of a specific speaker and then encode them into a vector and repeat it along timesteps. In this case, the reference works as a global condition throughout frames. A global trainable weight $W$ is then utilized to fuse the encoder outputs. We constrain $W$ via $\sum_{i}W_{i}=1$, forcing the network to find suitable weights for the encoder outputs. The decoder is designed to generate landmarks with transposed convolutions. Face landmarks and lip landmarks are modelled separately and then concatenated. To make sure the generated landmarks are smooth in timing and locations, we apply convolutions on these axis respectively.

\subsubsection{Training.} We randomly sample a frame of video and feed the corresponding landmarks into pose landmarks encoder to generate pose embeddings. Furthermore, we select a corresponding temporal window of mel-spectrogram as the input to mel-spectrogram encoder \cite{prajwal2020lip}. For the reference landmarks encoder, we utilize convolution blocks to extract features for every single frame of landmarks. We average these features and finally achieve an embedding vector. This vector is treated as a global condition that guides the decoder to generate better landmarks. Note that averaging these features enables us to use various frames of landmarks as a reference. Specifically, our model can accept only one frame of landmarks as a reference under limited conditions. All these three embeddings are then fused via weighted average manner for predicting lip movements. The landmark generator $G_l$ is simply trained with L1 distance:
\begin{equation}
    L_{G_l} = \frac{1}{NT}\sum_{i=1}^{N}\sum_{t=1}^{T} \lvert{l^a_i(t) - \hat{l}^a_i(t)}\rvert
\end{equation}

\subsection{Few-shot Landmark Translation}
\begin{figure}[t]
\begin{center}
   \includegraphics[width=1.0\linewidth]{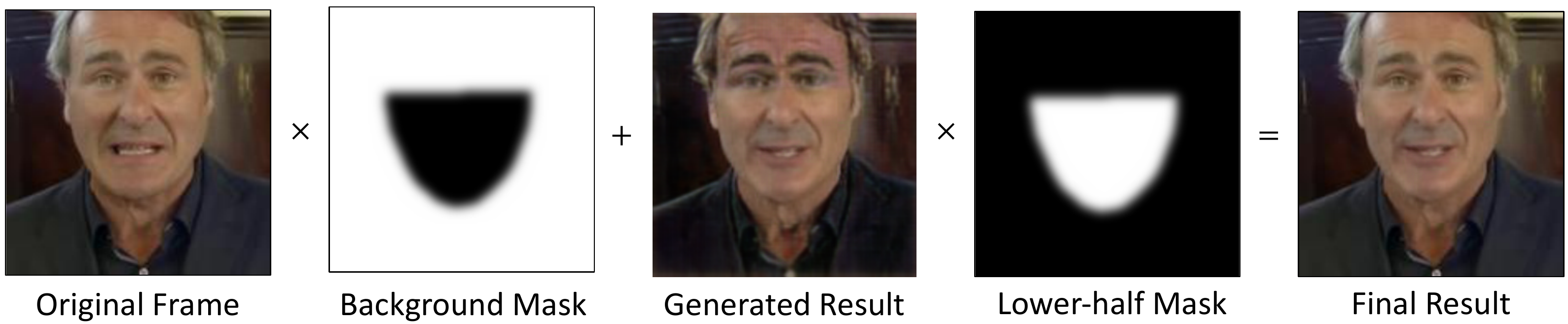}
\end{center}
   \caption{Composite the original frame with the generated result to get the final result.}
\label{fig:composite}
\end{figure}

The objective of this stage is to edit raw video content and change the lip movements to match the predicted landmarks $\hat{l}^a_i(t)$. To this end, we first mask out the lower-half face region and draw the predicted landmarks. This results in a sequence of composites $x_i(t)$, which acts as the pose and expression prior. Besides, we also input a reference frame $v_i(t')$ from which we use an embedder $E$ to extract personalized appearance features $e_i(t')$. Then, with $x_i(t)$ and $e_i(t')$, an image translation generator $G_v$ is adopted to synthesize the realistic lower-half face $\hat{v_i}(t)$. Finally, the edited frame $y_i(t)$ is obtained by compositing the generated lower-half face and the original frame (see Figure \ref{fig:composite}). Generally, the process of this stage can be formulated as follows:
\begin{equation}
    y_i(t) = G_v(x_i(t), E(v_i(t'))
\end{equation}
Moreover, our approach also supports feeding another speaker's reference image to transfer his/her appearance features to the target speaker.
 
\subsubsection{Network Architecture.} Our few-shot landmark translation stage consists of four modules: an appearance embedder $E$, a landmark translation generator $G_v$, and two discriminators $D_{vs}, D_{vt}$. The embedder projects a reference image to an appearance vector $e_i(t')$, which represents deep semantics of appearance texture. The generator takes a U-Net structure \cite{ronneberger2015u} with several AdaIN Resblocks. The AdaIN Resblocks is the same with the classical Resblock except the normalization layer. Through a learnable dense layer that calculates the weighted sum of speaker-specific embedding $e_i(t')$, the AdaIN Resblock get the affine parameters $\beta$ and $\gamma$ for instance normalization, where the appearance information is introduced. Note we only use the AdaIN Resblock in the smallest feature map size for the sake of manipulating only the deep semantics, for instance, whether the person has beard or not. Considering our situation is also analogous to image inpainting, we replace the vanilla convolution with gated convolution \cite{yu2019free} in the downsampling and upsampling layers, which helps improve the quality of synthesized results and promotes faster training convergence. Moreover, for better temporal coherence, our landmark translation generator takes a space-time volumes of conditional images $\{x_i(j)\vert j=t-p,...,t+p\}$ as input. For high-fidelity synthesis, we design two discriminators used for the GAN training. $D_{vs}$ is a per-frame spatial patch-based discriminator \cite{wang2018high}, and tries to distinguish real images from generated images. By contrast, $D_{vt}$ is a temporal patch-based discriminator and can judge based on temporal coherence. For its temporal input, we simply concatenate consecutive frames along the channel dimension. The structures of two discriminators are the same except the input size.

\subsubsection{Training.} Following the idea of \cite{zakharov2019few}, the training of few-shot landmark translation is divided into two phases: the meta-training phase and the fine-tuning phase. We train all the four networks in the meta-training phase while fix the embedder in the fine-tuning phase. The training objective is identical for both phases:
\begin{equation}
\label{equ:overall loss}
    L_v = L_{r} + \lambda_s * L_s + \lambda_t * L_t
\end{equation}
Here, the reconstruction loss $L_r$ measures the distance between the ground-truth image and the generated, using the perceptual loss \cite{johnson2016perceptual} calculated with VGG \cite{simonyan2014very} and VGGFace \cite{parkhi2015deep}. The adversarial loss terms, $L_s$ and $L_t$, are previously used in \cite{wang2018high}, which compromise GAN feature matching loss and patch GAN loss.

During the meta-training phase, our landmark translation model needs to be trained over a multi-speaker dataset which contains video footage of various speakers. After meta-training converges, our landmark translation model is able to synthesize realistic talking faces even for unseen speakers. Yet, there are might be some identity gaps between the generated and the ground-truth. A following fine-tuning phase can help bridge the gap. In the phase of fine-tuning, we fix parameters of the embedder $E$ and train the rest of networks. This phase proceeds on a single video footage and enables the generator to capture personalized facial details.

\section{Experiments}
In this section, we comprehensively compare our methods with against several SOTA methods in few-shot lip-synchronization as well as few-shot speech-driven talking head generation. We further verify the effectiveness of our ability to utilize heterogeneous data. Plus, ablation study on model components is also provided.

\subsection{Experiment Setting}
Two datasets of talking head videos are used for evalution: LRS2 \cite{afouras2018deep} (160p videos at 25 fps) and LRW \cite{Chung16lip} (256p videos at 25 fps). 

\begin{table*}
\begin{center}
\setlength{\tabcolsep}{4mm}
\begin{tabular}{|l|c|ccccc|c|}
\hline
\multirow{2}*{\textbf{Method}} & \multirow{2}*{\textbf{Dataset}} &\multicolumn{5}{c|}{\textbf{paired}} & \textbf{unpaired} \\
\cline{3-8}
~ & ~ & \textbf{FID} $\downarrow$ & \textbf{NME} $\downarrow$ & \textbf{SSIM} $\uparrow$ & \textbf{PSNR} $\uparrow$ & \textbf{LPIPS} $\downarrow$ & \textbf{FID} $\downarrow$ \\
\hline
LipGAN \cite{kr2019towards} & \multirow{3}*{LRS2} &10.8727 & 0.0160 & 0.9260 & 32.50 & 0.0543 & 10.8120\\
Wav2Lip \cite{prajwal2020lip} & ~ &8.6146 & 0.0163 & 0.9246 & 31.84 & 0.0624 & 12.8225\\
Ours & ~ &\textbf{4.9080} & \textbf{0.0136} & \textbf{0.9298} & \textbf{33.08} & \textbf{0.0414} & \textbf{9.4068}\\
\hline
LipGAN \cite{kr2019towards} & \multirow{3}*{LRW} & 8.2843 & \textbf{0.0090} & 0.9367 & 32.74 & 0.0377 & 8.3317\\
Wav2Lip \cite{prajwal2020lip} & ~   &5.0358 & 0.0099 & 0.9347 & 32.30 & 0.0465 & \textbf{6.1876}\\
Ours & ~    & \textbf{4.2515} & 0.0098 & \textbf{0.9372} & \textbf{32.79} & \textbf{0.0273} & 6.3544\\

\hline
\end{tabular}
\end{center}
\caption{We compare our method with fine-tuning with few-shot lip-synchronization methods on the sampled speakers from two datasets under paired and unpaired audio-visual settings, respectively. In the paired setting, we use the original audio-visual pair of videos while in the unpaired setting, the audio is substituted by the speech from other videos.}
\label{tab: quantitative comparison}
\end{table*}

\subsubsection{Baselines.}
For few-shot lip-synchronization methods, we select Wav2Lip \cite{prajwal2020lip} and LipGAN \cite{kr2019towards} as baselines, whose objective is essentially the same with ours. As for few-shot speech-driven talking head generation methods, we compare our method with MakeItTalk \cite{zhou2020makeittalk} and ATVGnet \cite{chen2019hierarchical}, which aim to animate static talking head images according to the audio input. 

\subsubsection{Metric.} In this paper, multiple metrics are adopted to evaluate the visual quality of the synthesized results as well as the lip synchronization. Structured similarity (SSIM) \cite{wang2004image}, Peak Signal-to-Noise Ratio (PSNR), and Learned Perceptual Image Patch Similarity (LPIPS) \cite{zhang2018unreasonable} are employed to measure the similarity between the generated image and the ground-truth image from the aspects of the global structure as well as the semantic structure. As for evaluating lip syncrhonization, we adopt the Normalized Mean Error (NME) \cite{yan2018survey} of landmarks extracted from generated videos, which is calculated as follows:
$$
    NME(l_t,\hat{l}_t) = \frac{1}{N}\sum_n \frac{\| l_{t,n} - \hat{l}_{t,n} \|_2}{d_t^{*}}
$$
where $l_{t,n}$ represents $n$-th keypoint of landmark at frame $t$ and we choose the bounding box diagonal as $d_t^{*}$. In the absence of ground-truth dubbing videos, we report the results based on Fr\'echet Inception Distance (FID) \cite{heusel2017gans}. Following the idea of \cite{wang2018video}, we develop the original FID metric to measure not only visual quality but also temporal consistency.

Apart from the effectiveness of the metric stated above, it is crucial to obtain feedbacks from human evaluators since people are sensitive to nuances in the audio-visual synchronization. Therefore, we ask evaluators to grade the shown visually dubbed videos according to three aspects: (1) overall visual quality, which involves authenticity, definition and fluency; (2) the consistency of speech and lip; (3) the identity similarity between the input video and the generated result. All of them are scored on the scale from 1 to 5 and the higher value represents better performance. The corresponding evaluation criteria is provided in the supplementary material.

\subsection{Quantitative Comparison}

\begin{table}
\begin{center}
\setlength{\tabcolsep}{1mm}
\begin{tabular}{lccc}
\hline
\textbf{Method} & \textbf{Visual Qual.} & \textbf{Lip Sync.} & \textbf{Identity Sim.}\\
\hline
ATVGnet \cite{chen2019hierarchical} & 2.049 & 2.649 & 2.533 \\
MakeItTalk \cite{zhou2020makeittalk}    & 3.091 & 2.195 & 3.332 \\
LipGAN \cite{kr2019towards} & 2.777 & 2.566 & 3.434 \\
Wav2Lip \cite{prajwal2020lip} & 3.159   & \textbf{3.516} & 3.576\\
Ours (w/o fine-tuning) & 3.736 & 3.430 & 3.700  \\
Ours (w/ fine-tuning) & \textbf{3.782} & 3.452 & \textbf{3.945} \\
\hline
\end{tabular}
\end{center}
\caption{Real-world evaluation on generated videos using unpaired data from the LRS2 test set. Evaluators grade the videos based on its visual quality, the degree of audio-lip synchronization and the identity similarity compare to the input, on a scale from 1 to 5.}
\label{tab: real-world evalution}
\end{table}

\begin{figure}[t]
\begin{center}
   \includegraphics[width=1.0\linewidth]{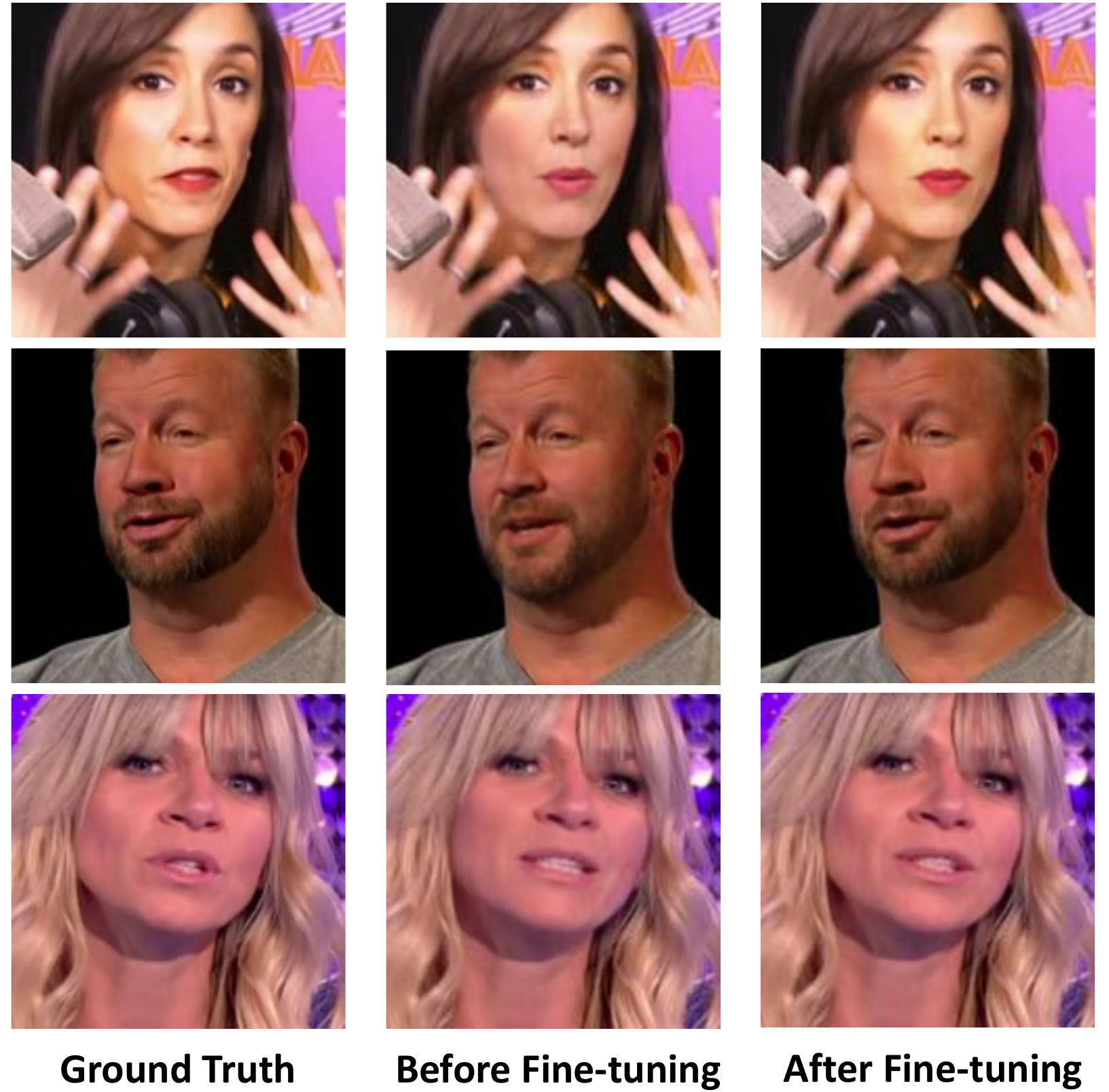}
\end{center}
   \caption{Comparison of synthesized results before and after fine-tuning. We show that before fine-tuning, our model is still able to generate realistic results.}
\label{fig:finetuning comparison}
\end{figure}

\begin{figure*}[t]
\begin{center}
\includegraphics[width=1.0\linewidth]{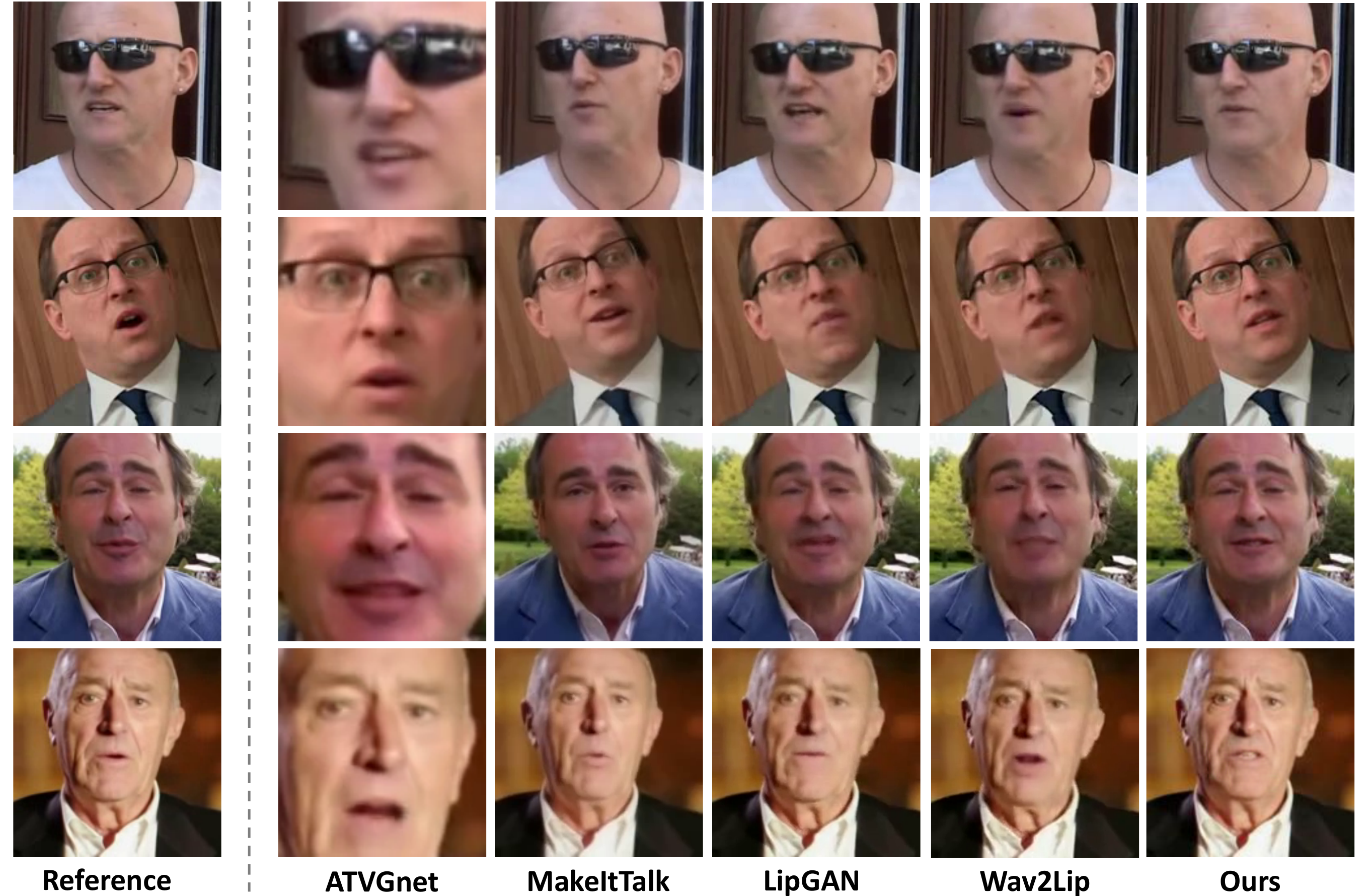}
\end{center}
   \caption{Visual dubbing results of our model (fine-tuned) and baselines on the LRS2 dataset. Synthesized results of our method are more realistic compared to baselines, especially the region around the mouth, and our model restores the facial features of the reference frame better.}
\label{fig:visual comparison 1}
\end{figure*}

In quantitative comparison, we focus on comparison with few-shot lip-synchronization. Since the task of MakeItTalk \cite{zhou2020makeittalk} and ATVGnet \cite{chen2019hierarchical} is to generate a full video given a static image instead of editing, their results may differ a lot from input talking head videos and thus it is unfair to compare our method with them when using comparison-based objective metrics. The quantitative comparisons are performed under two different settings. In the first case, the models take original audio-visual pairs of videos as the input, so that the corresponding ground-truth videos are available. As for the second setting, we replace the original audio channel with others, which is the common case of visual dubbing and we call it the unpaired case. In other words, for $v_i(t)$ and $a_{i'}(\tau)$, we set $i=i'$ in the paired setting while $i\neq i'$ in the unpaired setting. 

Since it will take tremendous time cost to fine-tune our landmark translation model for every single speaker in the LRS2 and LRW test set, we randomly sample 45 speakers from them respectively for the quantitative comparison study. Under the unpaired setting, for each selected speaker in LRS2 test set, we randomly sample 30 pieces of speech from other videos used for visual dubbing while 10 pieces for each selected speaker in LRW test set.

Table \ref{tab: quantitative comparison} presents the comparison results of paired and unpaired settings, where we use our model with fine-tuning for this comparison. SSIM, PSNR, LPIPS and NME are adopted when corresponding ground-truth videos are available, which is the case of the paired setting, but not applicable under the unpaired setting. As reported in the Table \ref{tab: quantitative comparison}, our fine-tuned model performs best on most metrics. 

\subsection{Real-world Evaluation}
In addition to the quantitative comparison between the ground-truth and generated videos, we also invite 80 human participants for subjective evaluation, where we focus on evaluating synthesized results in the unpaired setting and compare our method with all baselines. Similar to the unpaired setting in quantitative comparison, in the LRS2 test set, we sample 45 speakers and randomly select 30 pieces of speech for each of them. Thus 1.3k visually dubbed videos are synthesized for each method. During the evaluation, we show the synthesized video as well as one frame of the input video, and evaluators are asked to score the synthesized result according to the visual quality, audio-lip consistency and identity similarity. Note that each evaluator is asked to view and score the same amount of videos.

As we can see in Table \ref{tab: real-world evalution}, our method consistently outperforms baselines on visual quality and identity similarity, has competitive score of audio-lip synchronization compared to Wav2Lip. Specifically, our approach, whether is fine-tuned or not, obtains a significant improvement in visual quality and identity similarity, indicating more realistic and better identity preserving. For the audio-lip synchronization, the minuscule gap between the best baseline and ours also shows that our method has a competitive performance. Besides, we also demonstrate the availability of fine-tuning for our scheme, which improves the identity similarity a lot with minor affect to visual quality and audio-lip consistency. In contrast, fine-tuning is not applicable for other two few-shot lip-synchronization methods \cite{prajwal2020lip, kr2019towards} due to their end-to-end system, where fine-tuning may lead the model to overfit to one speaker's voice. We also present the generated results with and without fine-tuning in Figure \ref{fig:finetuning comparison}.

\subsection{Ablation Study}
\subsubsection{Training Data Ablation.}
The proposed framework enables us to fully make use of heterogeneous data. To verify the advantage of this feature, we conduct ablation study of training data on LRS2 dataset. Specifically, we first randomly select 50\% data from training set to train our two-stage network as the comparison base. In practical application, by adding additional data to landmark generation stage or landmark translation stage, we can further improve our model's performance. For instance, we can add blurry videos as training data for the first stage while GIF files for the second. For simplicity, we use the left 50\% data in LRS2 training set as the additional data and make a comparison experiment to discuss the effect of heterogeneous data. The comparison results are reported in Table \ref{tab: data ablation study}. Note that this NME is calculated using the predicted results of our landmark generator $G_l$. The ablation experiments show that by adding data to landmark generation model only, NME reduces a lot. Meanwhile, our landmark translation model obtains significant improvements on image-based metrics when more data is provided. These improvements demonstrate the capability of the proposed framework to leverage various video data. With such advantage, we can easily enlarge the training set using heterogeneous data, which enables a better generalization ability of the model. As for the similar results of landmark generators using different data on image-based metrics, we argue that the difference of landmarks drawn by landmark generators with different data ratio is actually very minor, especially in a low-resolution image, which finally leads to similar synthesized results.

\begin{table}[t]
\begin{center}
\begin{tabular}{cccccccc}
\hline
\multicolumn{2}{c}{\textbf{Data Ratio}} & \multirow{2}*{\textbf{FID} $\downarrow$} & \multirow{2}*{\textbf{SSIM} $\uparrow$} & \multirow{2}*{\textbf{PSNR} $\uparrow$ }& \multirow{2}*{\textbf{LPIPS} $\downarrow$} & \multirow{2}*{\textbf{NME} $\downarrow$}\\
\cline{1-2}
\textbf{\uppercase\expandafter{\romannumeral1}} & \textbf{\uppercase\expandafter{\romannumeral2}} & ~ & ~ & ~ & ~ & ~ \\
\hline
\multirow{2}*{50\%} &  50\% & 6.4850 & 0.9190 & 31.32 & 0.0488 & \multirow{2}*{0.0181}\\
~ &  100\% & 5.6012 & 0.9257 & 32.22 & 0.0440 & ~\\
\hline
\multirow{2}*{100\%} & 50\% & 6.5059 & 0.9194 & 31.33 & 0.0484 & \multirow{2}*{0.0071}\\
~ & 100\% & 5.5806 & 0.9257 & 32.19 & 0.0439 & ~\\
\hline
\end{tabular}
\end{center}
\caption{Ablation study of training data on LRS2 dataset using models w/ fine-tuning. \uppercase\expandafter{\romannumeral1}, \uppercase\expandafter{\romannumeral2} refer to the landmark generation stage and landmark translation stage respectively.}
\label{tab: data ablation study}
\end{table}

\subsubsection{Model Component Ablation.}
We also conduct ablation study of our landmark translation model, using the original audio-visual pair of videos in the LRS2 test set. Note we only compare the models without fine-tuning here. As reported in Table \ref{tab: component ablation study}, we can see that with the gated convolution, temporal discriminator and vggface loss, our model obtains significant improvement in the FID metric with a slight lift in other visual quality related metrics.

\begin{table}[t]
\begin{center}
\begin{tabular}{cccccccc}
\hline
\textbf{C. \uppercase\expandafter{\romannumeral1}} & \textbf{C. \uppercase\expandafter{\romannumeral2}} & \textbf{C. \uppercase\expandafter{\romannumeral3}} & \textbf{FID} $\downarrow$ & \textbf{SSIM} $\uparrow$ & \textbf{PSNR} $\uparrow$ & \textbf{LPIPS} $\downarrow$ \\
\hline
~ & \checkmark & \checkmark & 5.8130 & 0.9246 & 32.05 & 0.0447 \\ 
\checkmark & ~ & \checkmark & 6.2145 & 0.9255 & 32.09 & 0.0448 \\ 
\checkmark & \checkmark & ~ & 6.0960 & \textbf{0.9262} & 32.09 & 0.0448 \\ 
\checkmark & \checkmark & \checkmark & \textbf{5.5806} & 0.9257 & \textbf{32.19} & \textbf{0.0439} \\ 
\hline
\end{tabular}
\end{center}
\caption{Ablation study of our landmark translation model (w/o fine-tuning) on the LRS2 test set. Note that C. \uppercase\expandafter{\romannumeral1}, C. \uppercase\expandafter{\romannumeral2} and C. \uppercase\expandafter{\romannumeral3} denote gated convolution, temporal discriminator and vggface loss respectively.}
\label{tab: component ablation study}
\end{table}

\subsection{Qualitative Result}

\subsubsection{Visual Dubbing results.} The visual dubbing results of our fine-tuned model and the baselines are shown in Figure \ref{fig:visual comparison 1}. These results are synthesized by using the unpaired audio-visual data from the LRS2 test set, and thus there is no corresponding ground-truth videos. Compared to the baselines, our model synthesizes more verisimilar and sharp visual dubbing results, especially the region of the mouth and the teeth.

\begin{figure}[t]
\begin{center}
   \includegraphics[width=1.0\linewidth]{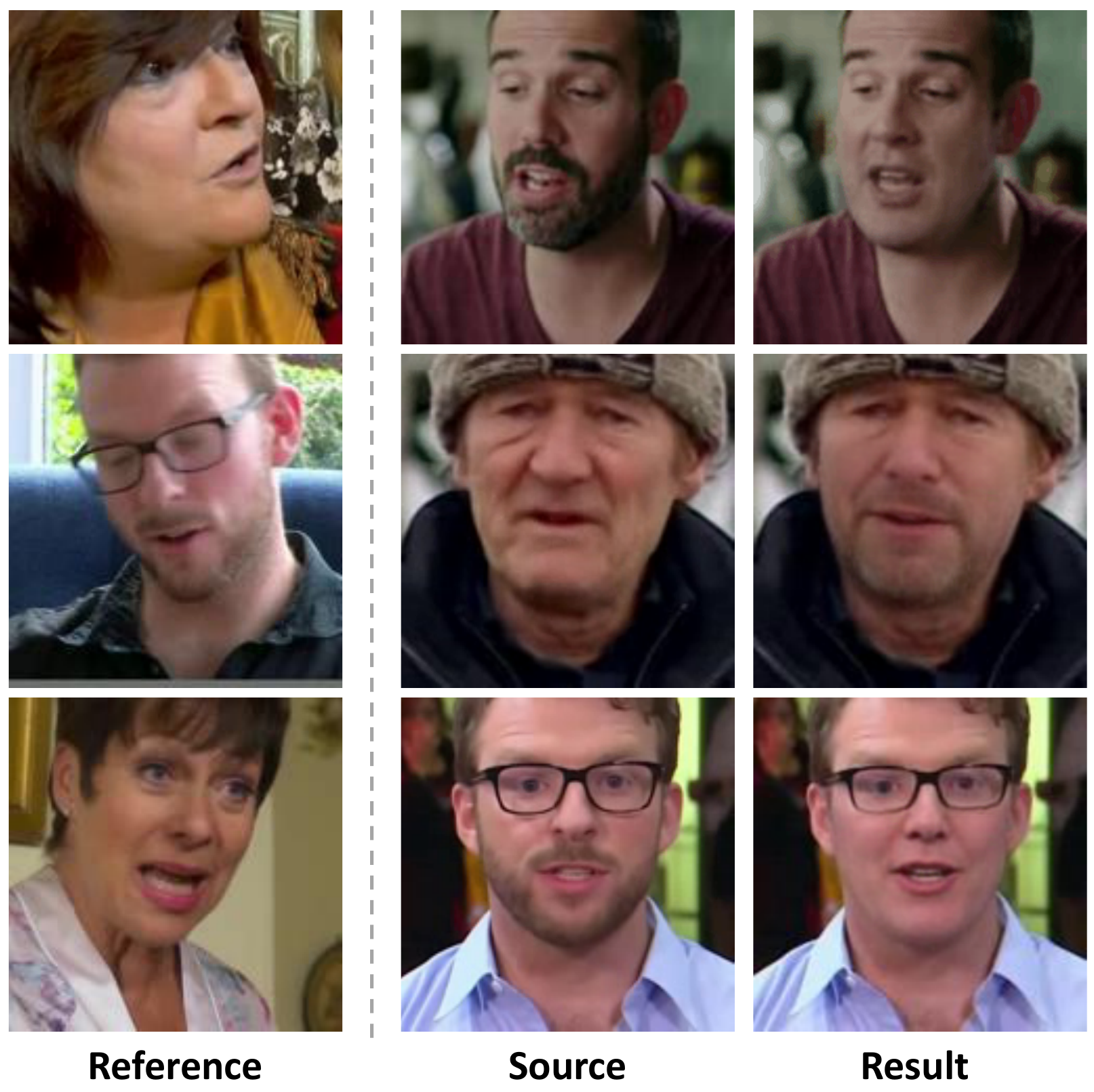}
\end{center}
   \caption{Examples of transferring lower-half face texture features. Provided a proper reference frame, our method is able to modify the appearance style of the target speaker, including mustache, wrinkle, lip color, etc. }
\label{fig:appearance transfer}
\end{figure}

\subsubsection{Appearance Transfer.}
\cite{huang2017arbitrary} employs AdaIN to perform style transferring. Similarly, our method is also able to transfer the appearance feature of the lower-half face, which is unavailable in previous visual dubbing approaches to the best of our knowledge. To achieve it, we can simply replace the reference frame $v_i(t')$ with a frame $v_j(t')$ from other videos in the landmark translation stage. Figure \ref{fig:appearance transfer} shows the appearance transferring results of our method. 

\section{Conclusion}
In this paper, we have proposed a novel two-stage framework to tackle the few-shot visual dubbing problem, which can use various video data flexibly. One stage is landmark generation and the other is landmark translation. Each stage performs an independent few-shot task and the landmark is employed as the intermediate representation. The proposed method can leverage heterogeneous data , which may be unsatisfactory in audios or pictures, and makes realistic visual dubbing possible under the few-shot setting. A further fine-tuning for the landmark translation stage is also enabled to solve the identity gap problem with no harm to the generalization ability on voice input. Moreover, our method supports appearance transferring by taking other's image as the reference frame. Extensive experiments prove that our proposed scheme is capable of generating highly realistic visual dubbing results.

\begin{acks}
The work was supported by the ByteDance Arnold and Data Platforms for the model training and subjective study. We would like to thank Jiani Chen, Han Feng, Shuangshuang He, Yuan Wan for their help in setting up the subjective evaluation. 
\end{acks}

\bibliographystyle{ACM-Reference-Format}
\balance
\bibliography{sample-base}

\end{document}